\documentclass{article}

\usepackage{PRIMEarxiv}

\usepackage[utf8]{inputenc} 
\usepackage[T1]{fontenc}    
\usepackage{hyperref}       
\usepackage{url}            
\usepackage{booktabs}       
\usepackage{amsfonts}       
\usepackage{nicefrac}       
\usepackage{microtype}      
\usepackage{lipsum}
\usepackage{fancyhdr}       
\usepackage{graphicx}       
\graphicspath{{media/}}     
\usepackage{amsmath}
\usepackage{amsthm}
\usepackage[caption=false]{subfig}
\usepackage{amsfonts}

\usepackage{multirow}

\usepackage{longtable}

\usepackage{booktabs}
\usepackage[load-configurations=version-1]{siunitx} 

\usepackage{wrapfig}
\usepackage{caption} 

\theoremstyle{definition}

\theoremstyle{plain}

\theoremstyle{remark}

\theoremstyle{definition}

\title{Detection of (Hidden) Emotions from Videos using Muscles Movements and Face Manifold Embedding
}

\author{
\quad\quad \quad \quad \quad Juni Kim \\
\quad \quad \quad \quad  \texttt{junickim@ohs.stanford.edu} \\
   \And
  \quad\quad\quad \quad \quad \quad \quad Zhikang Dong\\
   \quad\quad\quad \quad \quad \quad \quad Department of Applied Mathematics and Statistics \\
   \quad\quad\quad \quad \quad \quad \quad Stony Brook University \\ 
   \quad\quad \quad\quad \quad \quad \quad Stony Brook, NY, 11794, USA \\
  \quad \quad \quad\quad \quad \quad \quad \texttt{zhikang.dong.1@stonybrook.edu} \\
   \AND
  \quad \quad\quad Eric Guan \\
 \quad \quad \quad \texttt{ericguan2004@gmail.com} \\
   \And
  \quad  \quad \quad \quad \quad \quad \quad Judah Rosenthal \\
  \quad \quad \quad \quad \quad \quad \quad \texttt{judahrosenthal31@gmail.com} \\
      \AND
  \quad Shi Fu, Miriam Rafailovich \\
  \quad College of Engineering and Applied Sciences \\
  \quad Stony Brook University \\ 
  \quad Stony Brook, NY, 11794, USA \\
  \quad  \texttt{\{shi.fu, miriam.rafailovich\}@stonybrook.edu} \\
    \And
  Pawe\l \ Polak\thanks{Corresponding author. This report is a summary of the project done during the 2022 Garcia Summer Research Scholar Program for Gifted High School Students \url{https://www.stonybrook.edu/commcms/garcia/summer_program/program_description}.} \\
  Department of Applied Mathematics and Statistics \\
  Institute for Advanced Computational Science\\
  Stony Brook University \\ 
  Stony Brook, NY, 11794, USA \\
  \texttt{pawel.polak@stonybrook.edu} \\
}

\begin{document}

\maketitle

\begin{abstract}
We provide a new non-invasive, easy-to-scale for large amounts of subjects and a remotely accessible method for (hidden) emotion detection from videos of human faces. Our approach combines face manifold detection for accurate location of the face in the video with local face manifold embedding to create a common domain for the measurements of muscle micro-movements that is invariant to the movement of the subject in the video. In the next step, we employ the Digital Image Speckle Correlation (DISC) and the optical flow algorithm to compute the pattern of micro-movements in the face. The corresponding vector field is mapped back to the original space and superimposed on the original frames of the videos. Hence, the resulting videos include additional information about the direction of the movement of the muscles in the face. We take the publicly available CK++ dataset of visible emotions and add to it videos of the same format but with hidden emotions.
We process all the videos using our micro-movement detection and use the results to train a state-of-the-art network for emotions classification from videos---Frame Attention Network (FAN) from \cite{meng2019frame}. Although the original FAN model achieves very high out-of-sample performance on the original CK++ videos, it does not perform so well on hidden emotions videos. The performance improves significantly when the model is trained and tested on videos with the vector fields of muscle movements. Intuitively, the corresponding arrows serve as edges in the image that are easily captured by the convolutions filters in the FAN network.
\end{abstract}


\vspace{3mm}

\section{Introduction}\label{sec:Introduction}

Recognition of human faces has been an important topic in academia and industry for the past few decades \cite{goldstein1971identification,sirovich1987low,turk1991eigenfaces}. Although some great progress has been achieved, the biggest breakthrough came with the (geometric) deep learning revolution in recent years  (see \cite{bronstein2005three}). In addition to the detected face, it is important for the recent generations of deep models to detect also the facial expressions and classify human emotions.

Currently, there are deep learning models that utilize multi-frame attention mechanisms that already detect the facial expressions from videos and have high accuracy in identifying human feelings \cite{meng2019frame}. The CK++ dataset created in \cite{5543262} is a primary example of such videos. \cite{meng2019frame} construct a deep learning model that is trained and tested on the CK++ dataset and demonstrates very high out-of-sample accuracy. However, as shown by the results below, when the data includes videos with hidden emotions, where there are micro-movements in the facial muscles, the latest deep learning models such as \cite{meng2019frame} have much lower accuracy.



Hidden emotion detection is important because it can detect a person's subtle (potential subconscious) reaction to stimuli. Deep learning models that can accurately complete this task have broader applications in fields including medicine. For instance, they can provide better support for autistic people, who may have trouble showing emotions \cite{gong2013hidden}, as well as non-invasively detect emotional reactions from vegetative or comatose people \cite{sharon2013emotional}.

Other methods used to analyze the emotional reactions of the patients are EEG, EKG, and MRI. However, these are more invasive, expensive, and require specialized medical equipment and trained personnel. We propose a non-invasive, easy-to-scale for large amounts of patients, and remotely accessible alternative method of visible and hidden emotions detection using (potentially self-recorded) videos of the human face. Our approach can serve as a pre-diagnostic tool complementary to the aforementioned  medical-laboratory techniques, e.g., in the context of tele-medicine to analyze large amounts of remote patients.


Digital Image Speckle Correlation (DISC) was initially used for the non-contact measurement of material's mechanical properties and the detection of micro-movements on the surface of the material. Due to the traceable patterns of pores on the human skin, DISC has lately been used also in medicine to measure skin sample deformation \cite{Guanetal:04}; provide diagnostic and prognostic data for the management and treatment of vestibular schwannomas (acoustic neuroma) \cite{BhatnagarFioreRafailovichDavis:14}; and determine the optimal sites of injection for Botox \cite{Bhatnagaretal:13, Vermaetal:19}. Recently, we have used DISC to accurately analyse the facial muscle movement and classify the corresponding face expressions \cite{stonybrook, Wangetal:14, Pamudurthyetap:05} in a static environment from a short set of 2D images when the patient's head is not moving across frames. A method analogous to DISC, called Optical Flow, has been used in computer image analysis to track moving objects in a video. This method gives dynamic information about an objects movement and it has been widely used in deep neural networks (see \cite{ilg2017flownet} and references therein).

Once the face is in its canonical form we can apply DISC analysis and measure muscle movements using Optical Flow algorithms from the OpenCV library  \cite{opencv_library}. Once the optical flow results are obtained, we inverse-map the corresponding vectors onto the original frames of the video and create a new video with the original face enhanced by a vector field representing the muscle movements. Since many of the deep neural networks use convolutional layers that are particularly well suited for edges detection \cite{xie2015holistically}, our enhanced videos provide improved information for the deep neural networks to detect human emotions from the muscle movements depicted by our data processing pipeline. In order to demonstrate the improvements from the proposed method, we compare the performance of the Frame Attention Networks model from the aforementioned \cite{meng2019frame}, trained on the videos with and without our muscles movement analysis.

\vspace{-5mm}
\section{Features Construction and Experimental Results}\label{sec:Model} 
\vspace{-2mm}
We denote a gray scale video as a sequence of $p$ frames $\mathbf{V} = \{V_i\}_{i=1}^p$, where each of the frames $V_i \in \mathbb{R}^{\mathcal{N} \times \mathcal{M}}$  is a matrix of size corresponding to the resolution $\mathcal{N} \times \mathcal{M}$ of the video. 


\begin{figure}[h]
\centering
\includegraphics[width=.9\linewidth]{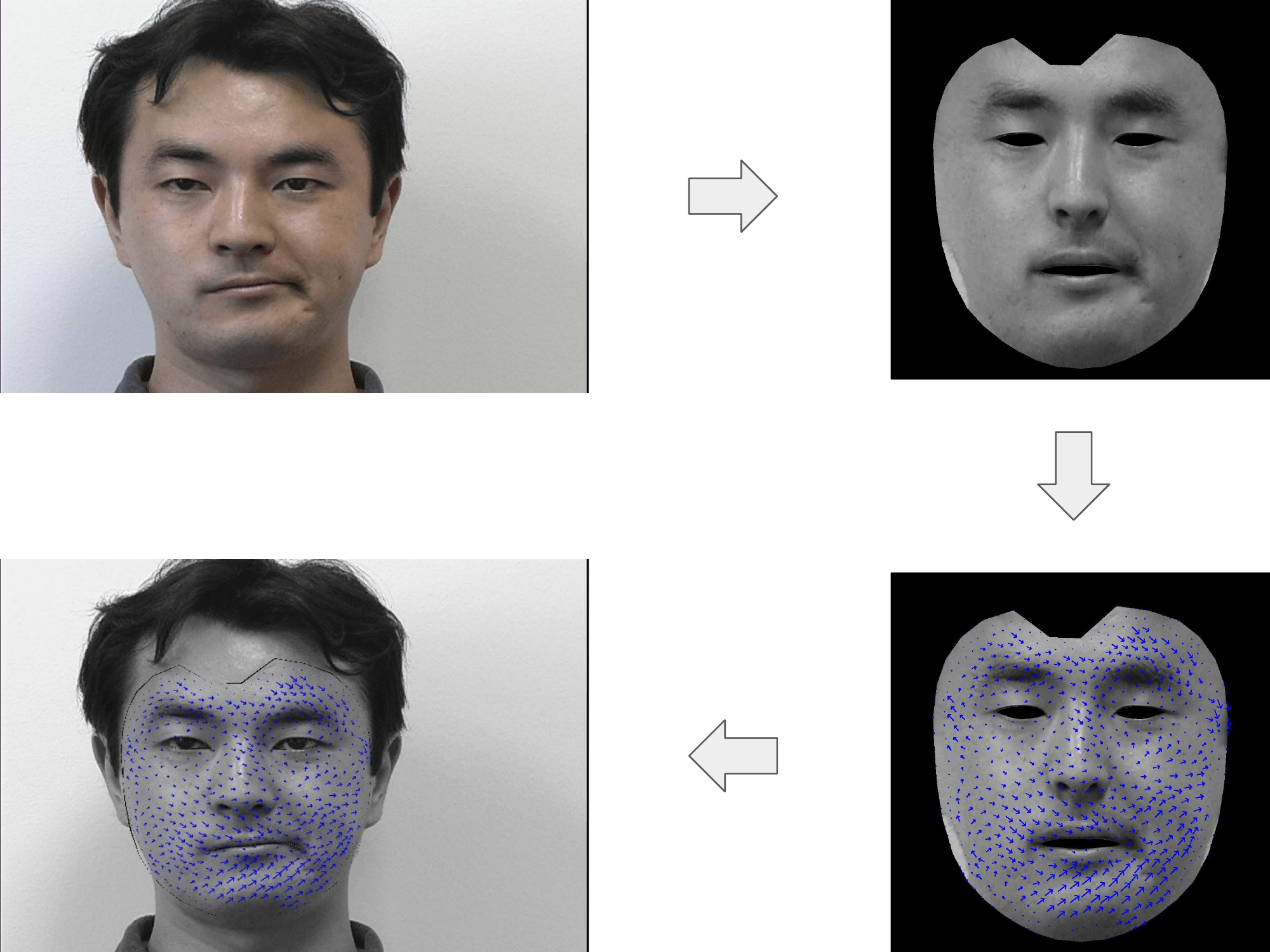}
\caption{Four steps of the analysis (Contempt).}
\label{fig:four_steps}
\end{figure}

Using Google's open-source MediaPipe library \cite{grishchenko2020attention}, we detect face manifolds and capture the triangulated face manifolds in the frames $F_i = \{(X_j^{(i)}, E_j^{(i)}) \}_{j=1}^\ell$ which form a graph  with $\ell$ landmarks $X_j^{(i)}$ and edges $E_j^{(i)}$ that connect the grid of landmarks, and provide $K$ triangles in each frame. These give us a sequence of face manifolds $\mathbf{F} = \{F_i\}_{i=1}^p$ for all of the frames in the video.

MediaPipe provides also a canonical face model $\widetilde{F} = \{(\widetilde{X}_j, \widetilde{E}_j) \}_{j=1}^\ell$ that acts as a flat surface representation of the face manifold. This canonical face model will serve as common canvas for our computations. Given the coordinates of the vertices $X_j^{(i)}$ in the frames of the video, and the coordinates of the corresponding vertices $\widetilde{X}_j$ in the canonical face model, we use triangle specific affine transformation \cite{croft2012unsolved} to map all the pixels in a particular triangle $\Delta_k$ to the corresponding triangle $\widetilde{\Delta}_k$ in the canonical face. 
Namely, for each pair of triangles $\Delta_k$ and $\widetilde{\Delta}_k$, we solve the 6 unknown parameters $m_{1k},m_{2k},\ldots,m_{6k}$ from the following bijective relationship

\begin{equation}\label{eq:affine}
\left[\begin{array}{c}
\widetilde{x}_{j} \\
\widetilde{y}_{j} \\ 
1
\end{array}\right]=\left[\begin{array}{lll}
m_{1k} & m_{2k} & m_{3k} \\
m_{4k} & m_{5k} & m_{6k} \\
0 & 0 & 1
\end{array}\right]\left[\begin{array}{c}
x_{j} \\
y_{j} \\
1
\end{array}\right]
\end{equation}
for $j=1,2,3$, where $(x_j, y_j)$ are the $x$- and $y$-coordinates of each vertex of $\Delta_k$.

Given the obtained linear mappings, we use the corresponding matrices as an affine transformation to map all the pixels inside the given triangle  $\Delta_k$  in the video frame to the corresponding triangle $\widetilde{\Delta}_k$ in the canonical frame. This way we obtain a sequence of canonical frames from the video $\widetilde{\mathbf{F}}=\{ \widetilde{F}_i\}_{i=1}^p$. These canonical embeddings allow us to measure the face muscles movements in the isolation from any movement of the subject in the video (up to the visibility of the face in the frames and accuracy of the face manifold computation). For the speed and scalability of the DISC computations, we use the optical flow method and the corresponding Lucas-Kanade algorithm \cite{lucas1981iterative}. 


Namely,  for each target pixel $q_1$, we have the neighbor pixels $q_2,q_3, \cdots, q_9$, together they consist of a $3\times3$ patch from the image. The local optical flow vector $\boldsymbol{d}$ will satisfy the equation

\begin{equation}
A^{T} A \boldsymbol{d}=A^{T} \boldsymbol{b},
\end{equation}
where 
$$
A=\left[\begin{array}{cc}
I_{x}\left(q_{1}\right) & I_{y}\left(q_{1}\right) \\
I_{x}\left(q_{2}\right) & I_{y}\left(q_{2}\right) \\
\vdots & \vdots \\
I_{x}\left(q_{9}\right) & I_{y}\left(q_{9}\right)
\end{array}\right] \quad \boldsymbol{d}=\left[\begin{array}{c}
d_{x} \\
d_{y}
\end{array}\right] \quad \boldsymbol{b}=\left[\begin{array}{c}
-I_{t}\left(q_{1}\right) \\
-I_{t}\left(q_{2}\right) \\
\vdots \\
-I_{t}\left(q_{9}\right)
\end{array}\right],
$$
$d_x$, $d_y$ are optical flow along $x$-axis and $y$-axis respectively, and $I_{x}\left(q_{j}\right), I_{y}\left(q_{j}\right), I_{t}\left(q_{j}\right)$ are the partial derivatives of the frame $I$ with respect to position $x, y$ and time $t$, evaluated at the point $q_{j}$.

Then it computes
$$
\left[\begin{array}{l}
d_{x} \\
d_{y}
\end{array}\right]=\left[\begin{array}{cc}
\sum_{i} I_{x}\left(q_{j}\right)^{2} & \sum_{i} I_{x}\left(q_{j}\right) I_{y}\left(q_{j}\right) \\
\sum_{i} I_{y}\left(q_{j}\right) I_{x}\left(q_{j}\right) & \sum_{i} I_{y}\left(q_{j}\right)^{2}
\end{array}\right]^{-1}\left[\begin{array}{c}
-\sum_{i} I_{x}\left(q_{j}\right) I_{t}\left(q_{j}\right) \\
-\sum_{i} I_{y}\left(q_{j}\right) I_{t}\left(q_{j}\right)
\end{array}\right].
$$

On the canonical face, $\forall (\widetilde{x}_j, \widetilde{y}_j) \in \widetilde{F}_j$ have the corresponding optical flow $
\left[\begin{array}{c}
\widetilde{x}_j + d_{\widetilde{x}_j} \\
\widetilde{y}_j + d_{\widetilde{y}_j}
\end{array}\right]$.

After computing the trajectories of muscles movements in the canonical faces, we use the inverses of our canonical embeddings to map the corresponding vector map back to the original frames from the video, creating a new video from these modified frames. Figure \ref{fig:four_steps} summarizes these four steps of our image analysis.

The resulting videos are used to train and test the FAN model from \cite{meng2019frame} and compared against the results of the same model trained on the original videos without the vectors representing the face muscles movements.

We also generate an extended dataset with 40 additional labeled videos, with similar length to the videos in the CK++ dataset, containing two subjects that are being exposed to positive and negative visual stimulation. In contrast to CK++, these new subjects are instructed not to express any emotions.\footnote{We confirmed after the recording that it is impossible to properly classify the emotions in the new videos using just the visual inspection.} Each video starts with a neutral frame of the subject before it was exposed to a visual stimulus, allowing us to use this first frame as a reference for computation (just like in CK++).


\begin{figure}[h]
\includegraphics[width=1\linewidth]{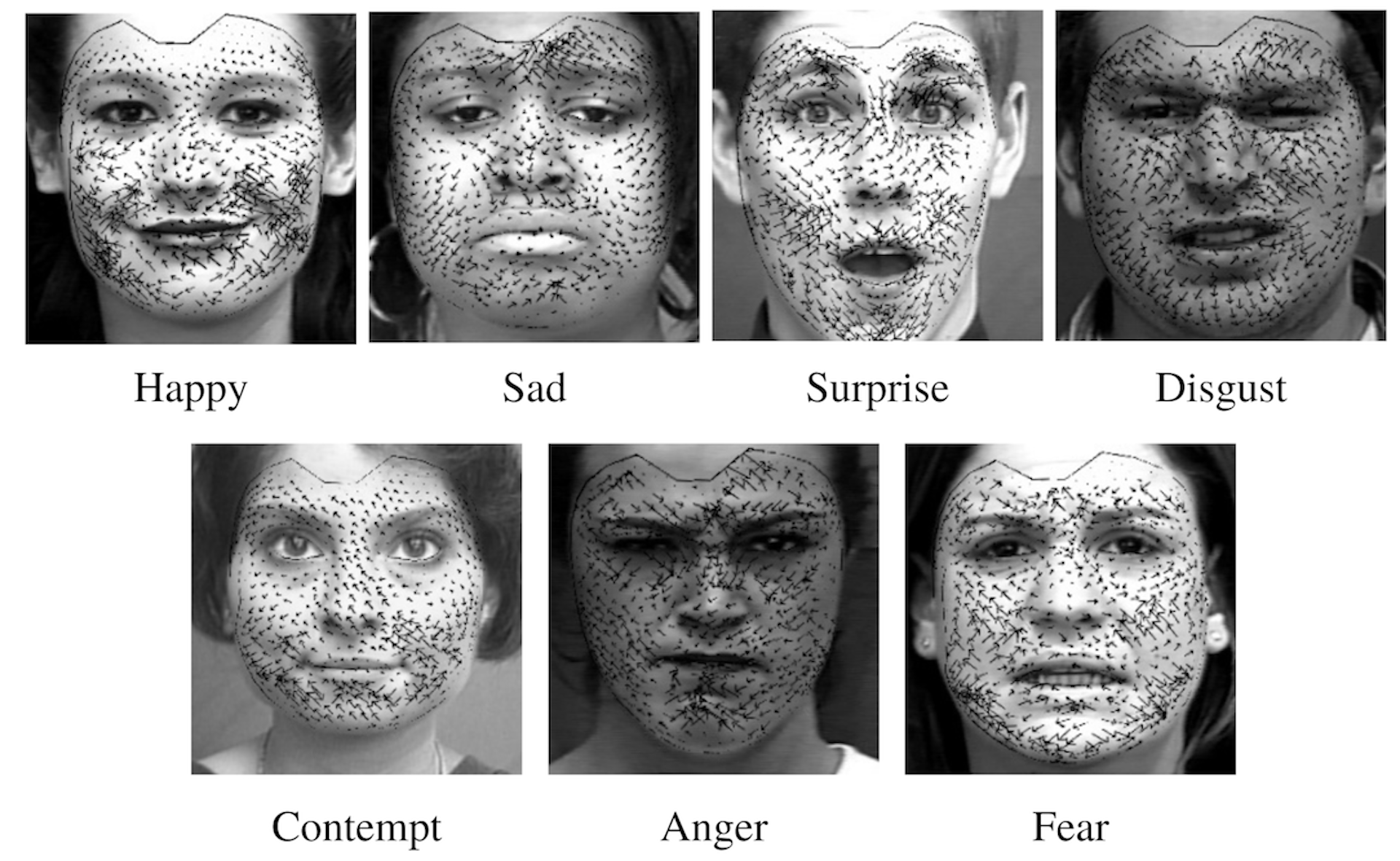}
\caption{Examples of frames from the CK++ dataset with the overlay of our muscles movements measurement.}
\label{fig:emotions}
\end{figure}

Different muscles movements are associated with different emotions, as shown in Figure \ref{fig:emotions}. The images used in the training of the model are of $224 \times 224$ dimension and the arrows take on average less than $10\%$ of the pixels in the frames. Hence, they do not cover any major part of the image or reduce other information that can be picked up by the model during the training.


We use the original and enhanced videos in a cross-validation exercise using the FAN model from \cite{meng2019frame}. Table \ref{tab:key_results} summarizes the out-of-sample accuracy results. The original FAN model with a pretrained ResNet network (trained on a large external set of images without DISC) performs almost perfectly on the original CK++ dataset and slightly better than our FAN+DISC model (results not included in Table \ref{tab:key_results}). We conjecture that during training the model did not consider the vectors field as significant information when emotions are obvious in videos. However, if we train the ResNet on the same data and/or include hidden emotions videos, the neural network in the FAN model recognizes the significance of the vector field. Out-of-sample accuracy is consequently significantly improved and the best model is FAN+DISC with pretrained ResNet.

\begin{table}[ht]
\begin{center}
\begin{tabular}{c| c c |c c | c c}
\multirow{3}{*}{Cross-ID} &\multicolumn{2}{c|}{CK++} & \multicolumn{4}{c}{CK++ \& Hidden Emotions}\\
& \multicolumn{2}{c|}{ResNet Trained} & \multicolumn{2}{c|}{ResNet Trained} & \multicolumn{2}{c}{Original ResNet Pretrained} \\ 
 & FAN  & FAN+DISC & FAN  & FAN+DISC& FAN & FAN+DISC\\ \hline
1 & 69.0 & \textbf{83.0} & 79.4 &  \textbf{84.1} & 92.1 &\textbf{92.6}\\
2 & 69.0 & \textbf{82.0}& \textbf{71.3} &  69.8 & 90.7 & \textbf{97.7}\\
3 & 67.0 & \textbf{82.0}& 70.5 &  \textbf{81.8} & \textbf{93.2} & \textbf{93.2}\\
4 & 77.0 & \textbf{82.0}& 81.2 &  \textbf{86.3} & 91.5& \textbf{95.7}\\
5 & 70.0 & \textbf{93.0} & 85.3 &  \textbf{89.2} & 99.0& \textbf{100}\\
6 & 72.0 & \textbf{81.0} & 74.8 &  \textbf{82.1} & 92.7& \textbf{100}\\
7 & 58.0 & \textbf{79.0} & 70.7 &  \textbf{82.8} & 90.9& \textbf{93.9}\\
8 & \textbf{82.0} & 75.0 & 76.8 &  \textbf{82.8} & 94.9& \textbf{97.0}\\
9 & 84.0 & \textbf{87.0} & 82.0 &  \textbf{84.7}& \textbf{99.1}& 97.3\\
10 & 77.0 & \textbf{86.0} & 84.6 &  \textbf{85.9}& \textbf{96.2}& 93.6\\
\hline
Avg & 72.5 & \textbf{83.0} & 77.6&  \textbf{83.0}& 94.0& \textbf{96.1}\\ \hline
\end{tabular}
\caption{Out-of-sample label predictions.}
\label{tab:key_results}
\end{center}
\end{table}
\section{Conclusions}\label{sec:Conclusions}
This short study shows the benefits of micro-movement detection for emotion classification. Our method creates additional features embedded into the original videos, so existing machine learning algorithms can be trained without any modification on micro-motions enhanced videos. The empirical results are very promising. In the next step, we plan to extend the dataset with more videos of hidden emotions, reduce the noise by smoothing the face manifold modeling over time, and further analyze and quantify micro-movements' role in hidden emotions videos classification. 

\newpage



\bibliographystyle{unsrt}  
\bibliography{references}

\end{document}